%% file: main.tex
\documentclass[10pt,twocolumn,letterpaper]{article}

 \usepackage{cvpr}              

\input{preamble}

\definecolor{cvprblue}{rgb}{0.21,0.49,0.74}
\usepackage[pagebackref,breaklinks,colorlinks,allcolors=cvprblue]{hyperref}

\def\confName{CVPR}
\def\confYear{2027}

\title{Signal Scaffolding: Unlocking the Potential of Lightweight Models for Self-Supervised Monocular Depth Estimation}

\author{Zhijie Wang, \ \   Li Zhang\thanks{Corresponding Author}, \ \   Yujie Chen,\ \  Shibiao Deng,\ \  Xin Gao,\ \  Houzhi Zhou, \ \  Jiangshuo Zhang\\
Department of Mathematics, Hefei University of Technology\\
Hefei, China\\
{\tt\small lizhang@hfut.edu.cn}
}

\begin{document}
\maketitle
\input{sec/0_abstract}

\input{sec/1_intro}
\input{sec/2_Related}

\input{sec/3_Method}

\input{sec/4_Experiments}

\input{sec/5_Conclusion}

\input{sec/LLM}
{
    \small
    \bibliographystyle{ieeenat_fullname}
    \bibliography{main}
}

 \input{sec/6_suppl}

\end{document}

%% file: preamble.tex
\usepackage{multirow}
\usepackage{pifont}
\usepackage{float}
\usepackage{bbding}
\usepackage{indentfirst}

\usepackage{latexsym}
\usepackage{amssymb}
\usepackage{amsmath}
\usepackage{amsthm}
\usepackage{booktabs}
\usepackage{enumitem}
\usepackage{graphicx}
\usepackage{color}
\usepackage{algorithmic}
\usepackage{textcomp} 
\usepackage{caption}



%% file: sec/0_abstract.tex
\begin{abstract}
\indent With the rising demand for edge devices, lightweight self-supervised monocular depth estimation methods have continued to evolve; however, existing methods generally suffer from poor accuracy. We find that the root cause may lies in the failure of conventional training regimes to fully exploit the inherent potential of lightweight models. To address this, we propose a novel three-phase training strategy for lightweight self-supervised monocular depth estimation, termed Signal Scaffolding (SS). Without introducing any additional dataset or parameters, we apply SS to ReForgeDepth, a lightweight network of our own design. Experiments on the KITTI and Cityscapes datasets demonstrate that our method achieves state-of-the-art performance in the accuracy-efficiency trade-off and running in real time on edge devices. These results suggest that lightweight models can be improved further within their existing capacity, and that the central question may be less about how parameters are allocated than about how to unlock this latent potential through training.
\end{abstract}

%% file: sec/1_intro.tex
\section{Introduction}

Monocular depth estimation is a core perception capability for autonomous driving and robotics. Supervised methods rely on costly LiDAR or manual annotation, whereas self-supervised methods recover depth from monocular video via photometric consistency alone \cite{godard2019digging, zhou2017un}. The pursuit of benchmark accuracy has driven a steady increase in model scale: transformer-based \cite{ranftl2021vision, zhao2022monovit} and diffusion-based self-supervised methods \cite{shao2024monodiffusion, ke2024repurposing, gui2025depthfm} routinely exceed tens of millions of parameters while pushing the state of the art. However, as perception tasks increasingly shift toward resource-constrained edge devices, deploying networks at such scale poses considerable challenges. Consequently, compact architectures have emerged as an active research direction \cite{litemono, zhang2025ledepth, zhu2026towards}.

\begin{figure}[htbp] 
  \includegraphics[width=0.5\textwidth]{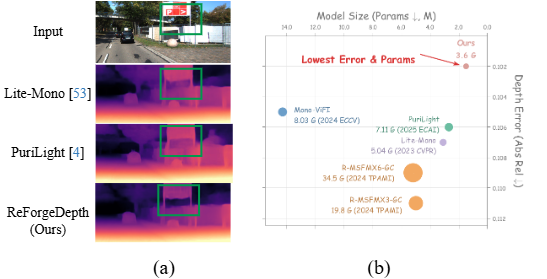} 
  \centering 
  \caption{Performance and qualitative comparison. \textbf{(a)} Qualitative depth predictions: the proposed ReForgeDepth recovers finer boundary details than representative lightweight methods while using the fewest parameters. \textbf{(b)} Accuracy-cost trade-off on KITTI, where ReForgeDepth sits closer to the upper right corner.}
  \label{fig:representative1} 
\end{figure}


Existing research on lightweight monocular depth estimation focuses primarily on network architecture design \cite{liu2026lea, zhu2026towards, gao2023edge, zhang2025lightweight}. For instance, Lyu \textit{et al.} \cite{lyu2021hr} introduced high-resolution skip connections to preserve spatial details; Zhang \textit{et al.} \cite{litemono} proposed a lightweight CNN-Transformer hybrid; and more recently, Chen \textit{et al.} \cite{chen2025purilight} introduced feature purification to improve estimation efficiency. However, despite continual architectural innovation, lightweight monocular depth estimation has long faced an unexplained accuracy ceiling: the most competitive compact models cluster within a narrow accuracy band, and pushing beyond it typically demands more parameters or more data \cite{liu2026lea, zhu2026towards, litemono, gao2023edge, zhang2025lightweight}.

In our experiments, we observe that pretraining does not always substantially improve the performance of lightweight networks—a phenomenon consistent with that reported by Muralidharan \textit{et al.} \cite{muralidharan2026pretraining} in supervised tasks. The prevailing view holds that this issue can only be resolved by further scaling up the network \cite{zhu2026towards} or by introducing additional semantic segmentation datasets \cite{wang2026mg}; yet both remedies run counter to the original intent of lightweight design and severely constrain the development of lightweight monocular depth estimation. We argue that the root cause might lie in the training pipeline inherited from networks of conventional scale: pretrain first, then fine-tune directly \cite{girshick2015fast, girshick2014rich}. Networks with rich redundancy naturally absorb pretraining knowledge \cite{he2019rethinking, kornblith2019better, yosinski2014transferable}; lightweight networks, with fewer parameters and lower redundancy, can likewise absorb this knowledge, yet ultimately transfer only a small fraction of it.

To address this challenging issue, we propose Signal Scaffolding (SS), a three-phase training strategy that extends the conventional pretrain-then-fine-tune pipeline. Its central insight is that, once network parameters and training data are fixed, the performance bottleneck is determined not solely by model capacity but also by the training strategy. We therefore opt to restructure the training pipeline so as to recover much of the representational capacity lost under the standard procedure. The strategy operates on how training is organized rather than on the parameters themselves. To instantiate and validate this strategy, we further design ReForgeDepth, a lightweight network that incorporates a Gated Multi-Dilation Block (GMD) and a Detail-Compensated Global Interaction (DCGI), together with an RGB Boundary Residual in the decoder, reducing the parameter count while leaving detail and receptive field intact. Applied to ReForgeDepth, SS leaves the architecture, parameters, and dataset untouched; the model still attains state-of-the-art performance in the accuracy-efficiency trade-off on KITTI and Cityscapes, and it also generalizes well to Make3D and runs in real time on edge devices. Our main contributions are summarized as follows:
\begin{itemize}
  \item We propose Signal Scaffolding (SS), a novel three-phase training strategy designed to recover much of the representational capacity lost under the standard training procedure; it substantially improves the accuracy of lightweight models without modifying the network architecture or introducing any additional parameters or dataset. \\
  \item  We design ReForgeDepth, a lightweight network whose core comprises a Gated Multi-Dilation Block (GMD) and a Detail-Compensated Global Interaction (DCGI), together with an RGB Boundary Residual in the decoder, reducing the parameter while leaving detail and receptive field intact. \\
  \item  Experiments on the KITTI and Cityscapes datasets demonstrate that ReForgeDepth, when trained with SS, achieves state-of-the-art performance in the accuracy-efficiency trade-off, and runs in real time on edge devices.
\end{itemize}


%% file: sec/2_Related.tex
\section{Related Work}	
\subsection{Monocular Depth Estimation.}

\indent Depth estimation is an ill-posed problem from a single image, so learning-based methods mainly differ in their source of supervision. Early models learned from dense annotations: Eigen \emph{et al.} \cite{eigen2014depth} established the CNN formulation, followed by up-projection \cite{laina2016deeper}, ordinal discretization \cite{fu2018deep}, and transformer backbones \cite{ranftl2021vision}. This accuracy depends on costly, range-limited, and noisy LiDAR or depth cameras, restricting data scale and label quality. Self-supervised methods instead use photometric consistency, judging depth by how well it reconstructs a nearby view. Godard \emph{et al.} \cite{godard2017} derived this objective from calibrated stereo with left--right consistency, while Zhou \emph{et al.} \cite{zhou2017un} showed that a moving monocular camera can jointly estimate depth and ego-motion. Because occlusions and static pixels corrupt photometric error, Monodepth2 \cite{godard2019digging} introduced minimum reprojection and auto-masking. In parallel, PackNet-SfM \cite{guizilini2020packnet} proposed 3D packing for resolution-robust generalization, and MonoViT \cite{zhao2022monovit} combined convolutional and transformer blocks for long-range context.

\subsection{Lightweight Depth Estimation}
Accurate depth networks are costly for edge devices, motivating self-supervised research. Early systems enabled embedded inference through compact encoder--decoder designs and pruning \cite{wofk2019fastdepth, elkerdawy2019lightweight}, followed by lightweight backbones \cite{poggi2018towards, poggi2022real} and hybrid CNN--Transformer architectures \cite{luginov2023swiftdepth, sui2024lightweight}. Lite-Mono \cite{litemono} showed that compact, sub-3M encoders can remain competitive. However, reduced capacity weakens fine structures and sharp depth discontinuities. This penalty has two sources: what a compact architecture cannot represent, and what it can represent but fails to reach under a direct self-supervised objective. The former is a genuine capacity limit; the latter is recoverable through the training signal alone, at no architectural cost---the source this paper targets.

\begin{figure*}[htbp] 
  \centering 
  \includegraphics[width=1.0\textwidth]{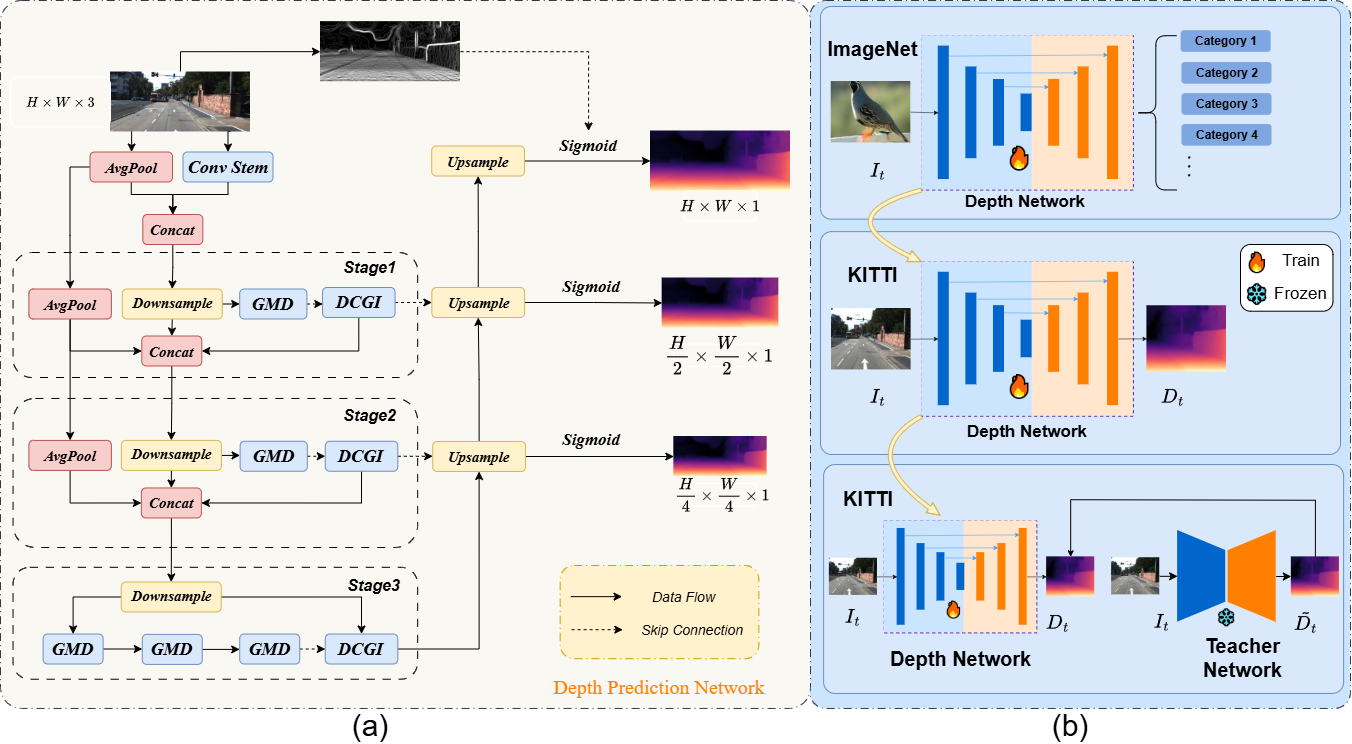} 
  \captionsetup{justification=centering}
  \caption{\textbf{(a)} Overall Architecture of the ReForgeDepth. The proposed architecture comprises three innovative modules: GMD, and DCGI, with detailed implementations illustrated in Figure {\color{red}\ref{fig:gmd-dcgi}}. \textbf{(b)} The proposed Signal Scaffolding training strategy.}
  \label{fig:ReForgeDepthMain}
\end{figure*}

\subsection{Training-Signal Design for Lightweight Depth Estimation}
\indent Lightweight depth estimation is often approached through architectural design, but the training signal also determines how effectively a compact model adapts to the target task. Prior work has explored teacher--student distillation and self-distillation for monocular depth estimation~\cite{wu2022toward,liu2023self,hu2023boosting,yu2025exploiting}, demonstrating the value of informative training signals. Conventional distillation, however, has the student mimic the teacher's outputs~\cite{hinton2015distilling}, with a ceiling set by the teacher; self-distillation~\cite{liu2023self} instead refines the model's own predictions. Both act as replacement signals rather than ones aligned with the student's representations. Signal Scaffolding (SS) departs from this paradigm: instead of allocating model capacity, it optimizes the training strategy of a fixed capacity---first adapting the student to the target domain through photometric self-supervision, then gradually introducing domain-matched depth guidance. This three-phase strategy aligns the guidance with the student's task-adapted representations, helping it better exploit its existing capacity.

%% file: sec/3_Method.tex
\section{Methods}


We propose ReForgeDepth, a lightweight architecture whose Gated Multi-Dilation Block and Detail-Compensated Global Interaction substantially reduce parameters,  enabling effective deployment on mobile devices (Figure~\ref{fig:ReForgeDepthMain}). The DepthNet adopts an encoder (Sec.~\ref{sec:DepthNet}) and decoder (Sec.~\ref{sec:Decoder}) design and is jointly optimized with a PoseNet (Sec.~\ref{sec:PoseNet}) in a self-supervised manner (Sec.~\ref{sec:Self-supervised}). A Signal Scaffolding (SS) training strategy (Sec.~\ref{sec:SS}) further yields sharper structural details and object contours without inflating the parameter budget or requiring additional data.

\begin{figure*}[htbp] 
  \centering 
  \includegraphics[width=0.9\textwidth]{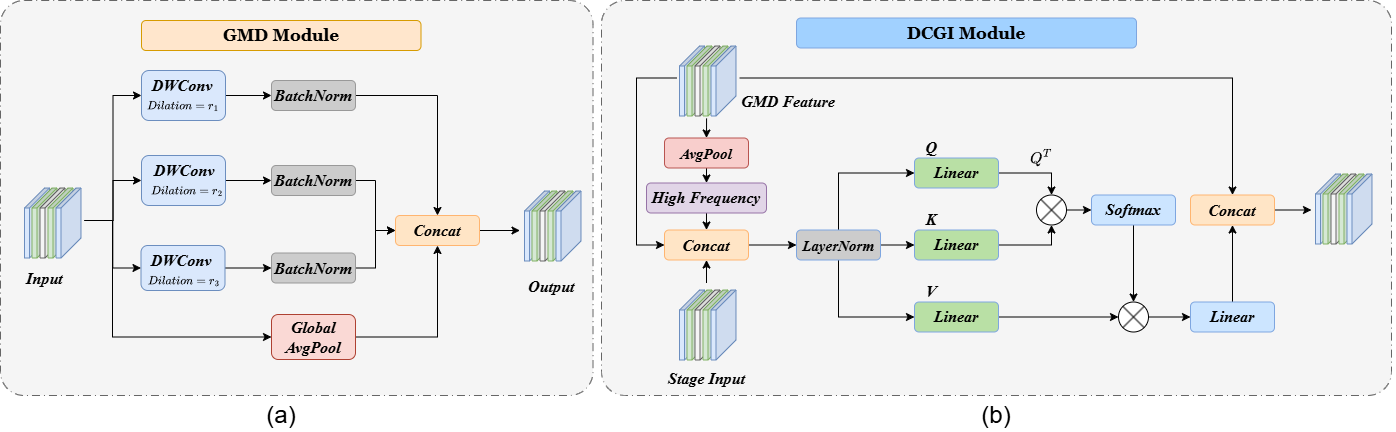} 
  \caption{Structures of the proposed GMD and DCGI module. }
  \label{fig:gmd-dcgi}
\end{figure*}

\subsection{Encoder Design}\label{sec:DepthNet}
\textbf{Motivation.}
A compact monocular-depth encoder must balance broad context for ambiguous regions with high-frequency cues for boundaries and thin structures. Stacked dilated blocks increase context at repeated feed-forward network (FFN) and projection cost, while fixed fusion cannot adapt to scene structure. Gated Multi-Dilation Block (GMD) therefore performs parallel consecutive dilated convolutions in one block with input-conditioned gates and road-oriented anisotropic kernels. At each stage end, Detail-Compensated Global Interaction (DCGI) couples local features with stage inputs, preserves signed high-frequency residuals, and adds global channel interaction. Together, GMD extracts adaptive local context, while DCGI restores detail and stage-level context efficiently.
\\
\indent \textbf{Gated Multi-Dilation Block (GMD).} GMD merges three consecutive dilated convolutions blocks, and their FFNs, into one residual block of parallel depthwise dilated convolutions with input-adaptive gates computed from the input (Figure~\ref{fig:gmd-dcgi}, (a)); branches use anisotropic $5\times3$ kernels whose taller axis favors the vertical structure dominant in road scenes. Given $X\in\mathbb{R}^{H\times W\times C}$:
\begin{equation}
Y_j=\operatorname{BN}\left(\operatorname{DWConv}_{5\times 3,,r_j}(X)\right),\qquad j=1,2,3,
\label{eq:pdb-branches}
\end{equation}
and the gating path
\begin{equation}
g=\sigma\left(\operatorname{PW}_{2}\left(\phi\left(\operatorname{PW}_{1}\left(\operatorname{GAP}(X)\right)\right)\right)\right)\in\mathbb{R}^{3C},
\label{eq:pdb-gate}
\end{equation}
where $\phi$ is ReLU, $\sigma$ the sigmoid, $\operatorname{GAP}$ is the Global Average Pooling operator and $\operatorname{PW}_{1,2}$ bias-free pointwise maps; the $3C$ outputs reshape into per-branch gates $g_j \in \mathbb{R}^C$, following the squeeze-and-excitation lineage \cite{hu2018squeeze,li2019selective} extended to dilation rates, and the zero-initialized gate head lets all branches contribute equally at the start of training. With $Y=\operatorname{ReLU}\left(\sum_{j=1}^{3} g_j\odot Y_j\right)$, the block output is
\begin{equation}
\hat{X}=X+\operatorname{DropPath}\left(\gamma\odot \operatorname{FFN}\left(\operatorname{LN}(Y)\right)\right),
\label{eq:pdb-fusion}
\end{equation}
where FFN is an MLP with GELU \cite{Dan2016GELU} and $\gamma$ is a LayerScale parameter \cite{touvron2021going}.
\\
\\
\indent \textbf{Detail-Compensated Global Interaction (DCGI).} Each stage terminates with a DCGI block (Figure~\ref{fig:gmd-dcgi}, (b)). The block takes the last local block's output $L=Z_i$ and the downsampled stage input $S=X_i$; a fixed, parameter-free pooling extracts the signed high-frequency residual $\hat{H}=L-\operatorname{AvgPool}_{3\times3}(L)$, the fine structure that lightweight networks lose first, kept unrectified so that both edge polarities survive. A depthwise--pointwise transform then refines it, and two learnable channel-wise scales form the attention input:
\begin{equation}
H=\operatorname{PW}\!\left(\operatorname{BN}\!\left(\operatorname{DWConv}(\hat{H})\right)\right), \ U=L+a\odot H+b\odot S.
\label{eq:fret-fusion}
\end{equation}
Attention is cross-covariance attention (XCA) \cite{ali2021xcit}, operating on channel cross-covariance at $O(C^2)$ cost: with $\hat{Q}$, $\hat{K}$ the $\ell_2$-normalized projections of $\operatorname{LN}(U+p)$ ($p$ a Fourier encoding), $V$ the value projection, $\tau$ a learnable per-head temperature, and $W_o$ the output projection,
\begin{equation}
A=\gamma_{\mathrm{attn}}\odot W_{o}\!\left(V\cdot\operatorname{Softmax} \!\left(\tau\odot\hat{Q}^{\top}\hat{K}\right)\right).
\label{eq:fret-xca}
\end{equation}
The block output is
\begin{equation}
\hat{Z}_{i}=L+\beta \cdot A+\operatorname{DropPath}\!\left(\gamma\odot \operatorname{FFN}\!\left(\operatorname{LN}(L+A)\right)\right),
\label{eq:fret-out}
\end{equation}
where $\gamma$, $\gamma_{\mathrm{attn}}$ are LayerScale parameters \cite{touvron2021going} and the scalar $\beta$ a zero-initialized gate on the direct attention residual \cite{bachlechner2021rezero}.

\subsection{Decoder and RGB Boundary Residual}
\label{sec:Decoder}

We keep the decoder deliberately simple: prior work \cite{litemono} shows a plain design already performs excellently in lightweight monocular depth, so we add no auxiliary components such as attention \cite{bae2022monoformer} and follow \cite{godard2019digging}: multi-scale features are fused via convolutional blocks and bilinearly upsampled, and a Sigmoid activation yields the final depth map at full resolution and $\frac{1}{2}$ and $\frac{1}{4}$ of it.

To sharpen boundaries, an RGB Boundary Residual (BR) augments the full-resolution output: the RGB image is converted to luminance and passed through fixed Sobel filters, giving the edge tensor $G=(G_x,G_y,\sqrt{G_x^2+G_y^2+\epsilon})$ ($\epsilon=10^{-6}$ for numerical stability). A depthwise--pointwise path encodes $G$ into an edge feature $E$, and the full-scale decoder feature $D$ is projected and upsampled to $\tilde{D}$:
\begin{equation}
E=\phi\!\left(\operatorname{PW}\!\left(\operatorname{DW}(G)\right)\right),\qquad
\tilde{D}=\operatorname{up}\!\left(\operatorname{PW}(D)\right),
\label{eq:boundary-enc}
\end{equation}
where $\operatorname{up}$ denotes upsampling and $\phi$ is GELU. The two are then concatenated and fused:
\begin{equation}
F=\phi\!\left(\operatorname{PW}\!\left(\operatorname{DW}\!\left(\operatorname{Concat}\!\left(E,\ \tilde{D}\right)\right)\right)\right).
\label{eq:boundary-fuse}
\end{equation}
A sigmoid gate modulates a residual projection, both single-channel:
\begin{equation}
\Delta_{B}=\sigma\!\left(W_{g}F\right)\odot W_{r}F
\label{eq:boundary-res}
\end{equation}
where $W_{g}$ and $W_{r}$ are independently learned $1\times1$
convolutions. With $u_{0}$ the base full-resolution logit:
\begin{equation}
\operatorname{disp}_{0}=\sigma\!\left(u_{0}+\Delta_{B}\right),\qquad
u_{0}=\operatorname{up}\!\left(\operatorname{Conv}_{3\times3}(D)\right).
\label{eq:boundary-out}
\end{equation}

\subsection{PoseNet}\label{sec:PoseNet}

\begin{figure}[ht] 
  \centering 
  \includegraphics[width=\linewidth]{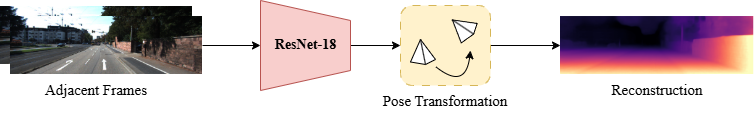}
  \captionsetup{justification=centering}
  \caption{PoseNet is employed to estimate the poses between adjacent frames. }
  \label{fig:posenet} 
\end{figure}

The pose network follows PoseNet \cite{zhou2017un, godard2019digging}: an ImageNet-pretrained ResNet-18 encoder takes two consecutive frames concatenated along the channel dimension, and a lightweight convolutional decoder predicts the 6-DoF relative pose (Figure \ref{fig:posenet}). This component is kept identical across all training configurations; it participates only in the self-supervised objective and is discarded at inference.

\subsection{Self-supervised Objective}\label{sec:Self-supervised}
Similar to \cite{zhou2017un},depth and camera motion are jointly optimized. For a target frame $I_t$, the depth network predicts $D_t$ and the pose network the relative transform $T_{t\to t'}$ for each neighboring source frame $I_{t'}$; source pixels gathered at the projected locations synthesize the target view,
\begin{equation}
{I}_{t'\to t}=I_{t'}\big\langle \operatorname{proj}(D_t,\, T_{t\to t'},\, K)\big\rangle,
\label{eq:warp}
\end{equation}
where $\langle\cdot\rangle$ is bilinear sampling and
$\operatorname{proj}(D_t, T_{t\to t'}, K)$ maps each target pixel to its
location in $I_{t'}$. Training minimizes the photometric discrepancy
\begin{equation}
pe(I_a,\,I_b)=\tfrac{\alpha}{2}\big(1-\operatorname{SSIM}(I_a,\, I_b)\big)+(1-\alpha)\,\|I_a-I_b\|_1,
\label{eq:pe}
\end{equation}
combining SSIM \cite{ssim} with the $L_1$ distance, as in \cite{zhou2017un, godard2019digging}, $\alpha=0.85$.

Occlusion and out-of-view pixels corrupt the signal: a pixel absent from a source frame admits no explaining warp, yet a loss summed over frames penalizes the network there. Following \cite{godard2019digging}, we take, per pixel, the minimum over source frames,
\begin{equation}
L_{p} = \min_{t'}\, pe\big(I_{t'} \big\langle \operatorname{proj}(D_{t}, T_{t\to{t'}}, K) \big\rangle, I_{t}\big),
\label{eq:minreproj}
\end{equation}
in place of the frame-summed loss.

When the camera is stationary, a surface lies at infinity, or an object moves rigidly with the camera, they are excluded by auto-masking \cite{godard2019digging},
\begin{equation}
    \begin{aligned}
      \mu = \big \lbrack \mathop{min}_{I_{s}\in[-1, 1]} L_{p}(I_{s}, I_{t})>\mathop{min}_{I_{s}\in[-1, 1]} L_{p}(I_{t'}, I_{t})  ~ \big \rbrack,
    \end{aligned}\label{eq:automask}
\end{equation}
where $[\cdot]$ is the Iverson bracket, the identity reprojection $I_{t\to t}$ warps the target to itself, and $\mu\in\{0,1\}$.

Since photometric error is uninformative in weakly textured regions, we additionally regularize the inverse depth with the edge-aware smoothness of \cite{godard2019digging},
\begin{equation}
L_{smooth}=\big|\partial_{x}d_{t}^{*}\big|\, e^{-|\partial_{x}I_t|} +\big|\partial_{y}d_{t}^{*}\big|\, e^{-|\partial_{y}I_t|},
\label{eq:smoothness}
\end{equation}
where $d_{t}^{*}=d_{t}/\bar{d}_{t}$ is the mean-normalized inverse depth. The base objective is
\begin{equation}
L_{\mathrm{base}}=\sum_{s=\{1,\frac{1}{2},\frac{1}{4}\}}\mu\odot L_{p} +\lambda\,L_{smooth},
\label{eq:base}
\end{equation}
where $\lambda=10^{-3}$ by default and $s$ denoted scale.

\begin{figure*}[htbp] 
\centering 
\includegraphics[width=0.96\textwidth]{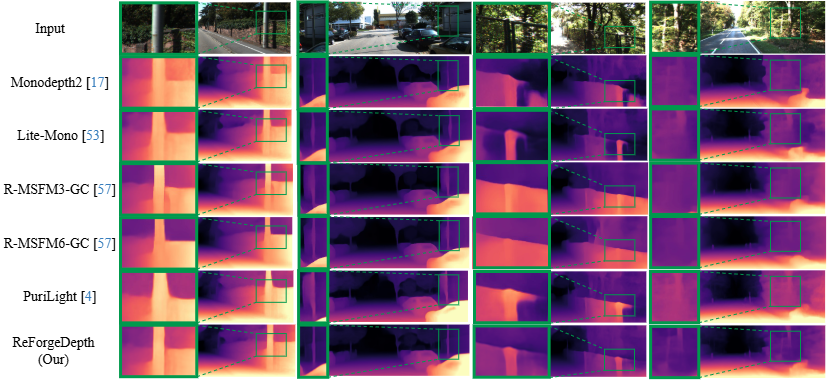} 
\captionsetup{justification=centering}
\caption{\textbf{Qualitative results on the KITTI Eigen test split.} Zoom-in views of the regions marked by colored boxes. Monodepth2~\cite{godard2019digging}  and Lite-Mono~\cite{litemono} blur object boundaries and lose fine details, whereas R-MSFM-GC~\cite{zhou2024recurrent} and PuriLight~\cite{chen2025purilight} oversmooth fine structures. ReForgeDepth, in contrast, preserves sharp boundaries and fine details.}
  \label{fig:example3} 
\end{figure*}

\subsection{Decoupling Depth from Capacity}
\label{sec:SS}

The performance plateau of lightweight depth models might stems from a capacity--training mismatch: direct fine-tuning may fail to reach the useful portion of a compact student's hypothesis space. Signal Scaffolding (SS) progressively organizes training signals, unlocking the student's recoverable capacity without architectural or data changes. A domain-matched teacher (Flex-Large \cite{zhu2026towards}) stimulates this latent capability. Our ablation experiments show that this potential does not come entirely from the teacher, but mainly from the inherent potential of the student model.

Both networks predict inverse depth up to an unknown per-image scale; we align each prediction by its median over the image domain $\Omega$,
\begin{equation}
 \widetilde{q}=\frac{q}{\operatorname{median}_{p\in\Omega}\,q(p)},
 \label{eq:median-normalization}
\end{equation}
where $q$ is the inverse depth obtained from the disparity $d\in(0,1)$. The aligned predictions are compared in the log domain, making the discrepancy a \emph{relative} error that large depths cannot dominate:
\begin{equation}
\begin{aligned}
 \ell_{\mathrm{KD}}^k&=\frac{1}{|\Omega_k|}\sum_{p\in\Omega_k}
       \sqrt{r_k(p)^2+\epsilon^2},\\
 r_k(p)&=\log\max\!\left(\widetilde{q}_{\mathrm{stu}}^k(p),\epsilon\right)
       -\log\max\!\left(\widetilde{q}_{\mathrm{tch}}^k(p),\epsilon\right),
\end{aligned}
\label{eq:kd-scale}
\end{equation}
where $k\in\{0,1,2\}$ indexes the full, $\tfrac{1}{2}$, and $\tfrac{1}{4}$ outputs, $\Omega_k$ the valid pixels at scale $k$, and $\epsilon=10^{-6}$.
The per-scale terms are aggregated with geometrically decaying weights $w_k=2^{-k}$:
\begin{equation}
 \mathcal{L}_{\mathrm{KD}}= \frac{\sum_{k=0}^{2}w_k\,\ell_{\mathrm{KD}}^k}{\sum_{k=0}^{2}w_k}.
 \label{eq:kd-total}
\end{equation}
The overall objective couples this with the self-supervised loss,
\begin{equation}
 \mathcal{L}=\mathcal{L}_{\mathrm{base}}+\lambda_e\,\mathcal{L}_{\mathrm{KD}},  \qquad \lambda_e=\eta\,\min\!\left(1,\frac{e}{5}\right),
 \label{eq:total-loss}
\end{equation}
where $\mathcal{L}_{\mathrm{base}}$ is the objective of Eq.~\eqref{eq:base}, $\eta=0.1$, and $e$ represents the number of training rounds counted from zero at third phase.

SS will not incur any additional parameters, data or inference overhead.

%% file: sec/4_Experiments.tex
\section{Experiments}
\label{sec:experiments}
This section presents ReForgeDepth and its Signal Scaffolding framework, and evaluates their effectiveness. For conciseness, the main text reports condensed experimental results, while the complete results are provided in the supplementary material.

\subsection{Implementation}
We evaluate ReForgeDepth in PyTorch on standard and cross-dataset benchmarks (Intel Core Ultra 7 270K CPU, NVIDIA RTX 5080, 48\,GB RAM). Training uses AdamW \cite{loshchilov2017decoupled} ($10^{-2}$ decay) and drop-path in GMD/DCGI: 60 epochs at $5\times10^{-4}$ (cosine learning rate schedule \cite{2017sgdr}) from scratch, or 30 at $1\times10^{-4}$ with ImageNet pretraining \cite{imagenet}, then 30 at $5\times10^{-5}$ with a frozen Flex-Large teacher.

\subsection{Datasets}

\noindent \textbf{KITTI.} We train and evaluate on KITTI \cite{kitti} with the standard Eigen split \cite{eig2015pr}: 39{,}180 monocular triplets $\{I_{t-1},I_t,I_{t+1}\}$ for training, 4{,}424 validation images, and 697 test images.
\\
\textbf{Cityscapes.} Cityscapes \cite{cordts2016cityscapes} features crowded urban streets with abundant moving objects and serves as our benchmark for dynamic scenes. We use the 58{,}335 training and 1{,}525 test images of \cite{feng2022disentangling} at a resolution of $416{\times}128$.
\\
\textbf{Make3D.} To assess cross-dataset generalization, we apply the KITTI-trained model directly to Make3D \cite{make3d}, which provides 134 outdoor test images, without any fine-tuning.

\subsection{Results}
We evaluate ReForgeDepth on the KITTI \cite{kitti} and Cityscapes \cite{cordts2016cityscapes} benchmarks (Tables~\ref{tab:kittiandcityscapes})\footnote{More qualitative results and visual comparisons are shown in the supplementary material and supplementary video.}. At KITTI 640$\times$192, with 1.5M parameters it attains Abs Rel 0.102, the best among lightweight methods---0.004 ahead of LEDepth \cite{zhang2025ledepth} and PuriLight \cite{chen2025purilight}---and surpasses Flex-Small \cite{zhu2026towards} with under 25\% of its parameters; only Flex-Medium/Large \cite{zhu2026towards} stay ahead. At 1024$\times$320 it again achieves the best Abs Rel, matching the best published result even without ImageNet pretraining. On Cityscapes it reaches 0.091, again the best among lightweight methods, surpassing all high-capacity baselines except Flex-Medium/Large \cite{zhu2026towards}. Across all settings, ReForgeDepth markedly shifts the accuracy--parameter trade-off beyond same-scale models.

Figure~\ref{fig:example3} compares ReForgeDepth with Monodepth2 \cite{godard2019digging} and recent lightweight models on KITTI \cite{kitti}: it renders sharper depth discontinuities at object boundaries, where the others over-smooth edges or leak depth, and recovers fine structure with fewer artifacts, consistent with Table~\ref{tab:kittiandcityscapes}. Transferred to Make3D \cite{make3d} without fine-tuning \cite{godard2019digging}, it outperforms all lightweight methods (Table~\ref{tab:make3d}) with clearer boundaries and cleaner geometry (See supplementary materials). Overall, ReForgeDepth delivers the best accuracy--efficiency trade-off: 1.5M parameters matching models 4--22$\times$ larger.

\begin{table*}[ht]
\centering
\captionsetup{justification=centering}
\caption{\textbf{Quantitative results.} We compare ReForgeDepth with representative methods on the KITTI Eigen split \cite{eig2015pr}. The best and second-best results in lightweight are highlighted in \textbf{bold} and \underline{underlined}, respectively. M: KITTI monocular videos.}
\footnotesize
\renewcommand{\arraystretch}{0.8}
\setlength{\tabcolsep}{3pt} 
\resizebox{\textwidth}{!}{%
\begin{tabular}{ccccccccccc}
\toprule
\multirow{2}{*}{Method} & \multirow{2}{*}{Year} & \multirow{2}{*}{Train} &
\multicolumn{4}{c}{\textbf{Depth Error} $\downarrow$} &
\multicolumn{3}{c}{\textbf{Depth Accuracy} $\uparrow$} &
\multirow{2}{*}{\begin{tabular}[c]{@{}c@{}} \textbf{Model Size} $\downarrow$\\ Params.\end{tabular}} \\
 & & & Abs Rel & Sq Rel & RMSE & RMSE log & $\delta < 1.25$ & $\delta < 1.25^2$ & $\delta < 1.25^3$ & \\ 
\midrule
\midrule
\multicolumn{11}{c}{\textit{\textbf{KITTI(640$\times$192)}}} \\
\multicolumn{11}{c}{\textit{Lightweight ($<$5M)}} \\
\midrule

Lite-Mono-tiny  \cite{litemono} & 2023    & M    & {0.110} & {0.837} & {4.710} & {0.187} & {0.880} & {0.960} & {0.982}  & {2.2M}  \\ 
Lite-Mono-small  \cite{litemono} & 2023    & M    & {0.110} & {0.802} & {4.671} & {0.186} & {0.879} & {0.961} & {0.982}  & {2.5M}  \\ 
Lite-Mono \cite{litemono}  & 2023    & M    & {0.107} & {0.765} & {4.561} & {0.183} & {0.886} & {0.963} & {0.983}  & {3.1M}  \\ 
MViTDepth-tiny \cite{gao2023edge}    & 2023    & M    & {0.111} & {0.825} & {4.731} & {0.187} & {0.881} & {0.961} & {0.983}  & \underline{1.8M}  \\ 
MViTDepth-small \cite{gao2023edge}    & 2023    & M    & {0.107} & {0.788} & {4.579} & {0.183} & {0.887} & {0.963} & {0.983}  & {2.8M}  \\ 
R-MSFM3-GC \cite{zhou2024recurrent}  & 2024 & M & 0.112   & 0.789  & 4.621 & 0.189    & 0.880 &  0.960 & 0.982 & 3.5M   \\ 
R-MSFM6-GC \cite{zhou2024recurrent}  & 2024  & M & 0.112   & 0.789  & 4.597  & 0.189    & 0.881 & 0.961 & 0.981 & 3.8M   \\ 
LEDepth \cite{zhang2025ledepth}  & 2025    & M    & \underline{0.106} & {0.780} & {4.602} & {0.183} &{0.888} & {0.963} & \textbf{0.985}  & {3.1M}  \\ 
ADDepth-tiny \cite{zhang2025lightweight}   & 2025    & M    & {0.112} & {0.815} & {4.642} & {0.186} & {0.880} & {0.962} & {0.983}  & \underline{1.8M}  \\ 
ADDepth-small \cite{zhang2025lightweight}   & 2025    & M    & {0.108} & \underline{0.741} & {4.503} & {0.182} & {0.885} & \textbf{0.965} & \underline{0.984}  & {2.8M}  \\ 
EdgeMono \cite{wang2025edgemono}   & 2025    & M    & {0.107} & {0.799} & {4.606} & {0.183} & {0.886} & {0.962} & {0.982}  & {4.3M}  \\ 
PuriLight \cite{chen2025purilight}  & 2025 & M  & \underline{0.106}   & {0.747}  & {4.536} & {0.182}    & \underline{0.890} & \underline{0.964}  & \underline{0.984} & {2.7M}   \\
LEA-depth-tiny \cite{liu2026lea}  & 2026    & M    & {0.109} & {0.776} & {4.528} & {0.182} & {0.884} & \underline{0.964} & {0.983}  & \underline{1.8M}  \\ 
LEA-depth-small \cite{liu2026lea}  & 2026    & M    & {0.107} & {0.758} & \underline{4.489} & \underline{0.181} & {0.888} & \underline{0.964} & \underline{0.984}  & {3.0M}  \\ 
Flex-Nano \cite{zhu2026towards} & 2026    & M    & {0.110} & {0.794} & {4.678} & {0.184} & {0.878} & {0.961} & {0.983}  & \textbf{1.5M}  \\ 
Ours & 2026  & M  & \textbf{0.102}   &\textbf{0.726}  & \textbf{4.459} & \textbf{0.179}    & \textbf{0.891} & \underline{0.964}  & {0.983} & \textbf{1.5M}   \\
\midrule
\multicolumn{11}{c}{\textit{\textbf{Cityscapes (416$\times$128 )}}} \\
\multicolumn{11}{c}{\textit{Lightweight ($<$5M)}} \\
\midrule

Flex-Nano \cite{zhu2026towards}  & 2026 & M & 0.107   & 1.261  & 6.133 & 0.164    & {0.893} &  0.971 & \underline{0.989} & \textbf{1.5M}   \\ 
PuriLight\cite{chen2025purilight} & 2025  & M & \underline{0.099}   & \underline{1.014}  & \underline{5.892}  & \underline{0.153}    & \underline{0.901} & \underline{0.976} & \textbf{0.992} & \underline{2.7M}   \\ 

Ours  & 2026  & M  & \textbf{0.091}   &\textbf{0.918}  & \textbf{5.590} & \textbf{0.146}    & \textbf{0.912} & \textbf{0.978}  & \textbf{0.992} & \textbf{1.5M}   \\

\bottomrule
\bottomrule
\end{tabular}%
}
\label{tab:kittiandcityscapes}
\end{table*}

\begin{table}[ht]
\centering
\captionsetup{justification=centering}
\caption{\textbf{Quantitative results.} Comparison of ReForgeDepth with representative methods on the Make3D \cite{make3d} dataset.}
\resizebox{0.45\textwidth}{!}{%
\begin{tabular}{cccccc}
\toprule
Method     & Abs Rel$\downarrow$ & Sq Rel$\downarrow$ & RMSE$\downarrow$  & RMSE log$\downarrow$ & Params$\downarrow$ \\ \midrule
DDVO \cite{wang2018}      & 0.387   & 4.720  & 8.090 & 0.204    & 28.1M  \\ 
Monodepth2 \cite{godard2019digging} & 0.322   & 3.589  & 7.417 & 0.163    & 14.3M  \\ 
R-MSFM6 \cite{rmsfm}    & 0.334   & 3.285  & 7.212 & 0.169    & 3.8M   \\ 
Lite-Mono-tiny \cite{litemono}  & 0.318   & 3.217  & 7.234 & 0.164    & 2.2M   \\ 
Flex-Small \cite{zhu2026towards}  & 0.315   & 3.127  & \textbf{6.865} & \textbf{0.159}    & 6.1M   \\ 
ReForgeDepth(Ours)      & \textbf{0.311}   & \textbf{3.094}  & {6.995} & {0.160}    & \textbf{1.5M}   \\ 
\bottomrule
\end{tabular}%
}
\label{tab:make3d}
\end{table}

\subsection{Complexity and Runtime}

Table~\ref{tab:complexity} reports model complexity and speed at 640$\times$192 on a MacBook Air (M2) and an iPhone 12 (A14), measured with Apple Core ML (Float32) and averaged over 200 runs after 50 warm-ups. ReForgeDepth attains the highest FPS on both devices, sustaining real-time inference on the iPhone 12 with the best accuracy--efficiency trade-off.

\noindent\textbf{Why FLOPs fail to predict on-device latency.} On mobile SoCs, latency depends more on operator-level hardware support than on FLOPs. Although Monodepth2-Res18 \cite{godard2019digging} requires 2.2$\times$ more FLOPs than ReForgeDepth, both achieve nearly identical latency on the iPhone 12 because Monodepth2-Res18's convolution-only architecture is well optimized by mobile inference engines. Conversely, PuriLight \cite{chen2025purilight} has modest FLOPs but runs significantly slower because its frequency-domain operations are poorly supported on mobile devices.

\begin{table}[ht]
\centering
\captionsetup{justification=centering}
\caption{\textbf{Model complexity and inference speed.} All models process 640$\times$192 inputs. Params: number of parameters; FLOPs:floating-point operations; FPS: frames per second.}
\resizebox{0.48\textwidth}{!}{%
\begin{tabular}{cccccccc}  
\toprule  
\multirow{2}{*}{Method} & \multicolumn{3}{c}{Full Model$\downarrow$}                 & \multicolumn{2}{c}{Macbook Air (M2)}    & \multicolumn{2}{c}{iPhone 12 (A14)}           \\ 
     & \multicolumn{1}{c}{Params(M)} & FLOPs(G) & Abs Rel & \multicolumn{1}{c}{Speed (ms) $\downarrow$} & FPS $\uparrow$ & \multicolumn{1}{c}{Speed (ms) $\downarrow$} & FPS $\uparrow$ \\  
\midrule
Monodepth2  \cite{godard2019digging}                &  \multicolumn{1}{c}{14.3}       & 8.0  &0.115  & \multicolumn{1}{c}{21.36}     & 47   & \multicolumn{1}{c}{30.47}     & 33        \\ 
R-MSFM3 \cite{rmsfm}                 &  \multicolumn{1}{c}{3.5}       & 16.5 & 0.114     & \multicolumn{1}{c}{21.61}      & 46    & \multicolumn{1}{c}{47.53}     & 21      \\ 
Lite-Mono \cite{litemono}            &  \multicolumn{1}{c}{3.1}       & 5.0 & 0.107    & \multicolumn{1}{c}{18.27}      & 55     & \multicolumn{1}{c}{41.65}     & 24     \\ 
PuriLight  \cite{chen2025purilight}        &  \multicolumn{1}{c}{{2.7}}       & {7.1}  & 0.106    & \multicolumn{1}{c}{{37.26}}      & {27}       & \multicolumn{1}{c}{60.64}     & 16   \\
ReForgeDepth        &  \multicolumn{1}{c}{\textbf{1.5}}       & \textbf{3.6}   & \textbf{0.102}    & \multicolumn{1}{c}{\textbf{17.28}}      & \textbf{58}      & \multicolumn{1}{c} {\textbf{29.80}}     & \textbf{34}    \\  
\bottomrule
\end{tabular}%
}
\label{tab:complexity}
\end{table}

\begin{table}[ht]
\centering
\caption{\textbf{Generalization of the Signal Scaffolding (SS) on the KITTI Eigen split.} ``w/o SS'' denotes training with ImageNet initialization only, and ``w/ SS'' further applies Signal Scaffolding.}
\resizebox{0.49\textwidth}{!}{%
\begin{tabular}{l cc cc cc}
\toprule
& \multicolumn{2}{c}{Abs Rel$\downarrow$} & \multicolumn{2}{c}{Sq Rel$\downarrow$} & \multicolumn{2}{c}{RMSE$\downarrow$}  \\
\cmidrule(lr){2-3} \cmidrule(lr){4-5} \cmidrule(lr){6-7}
Method & w/o & w/ & w/o & w/ & w/o & w/  \\
\midrule
\multicolumn{7}{l}{\textit{(a) Teacher: Lite-Mono-8M}} \\
Lite-Mono-tiny & 0.110 & 0.106 & 0.837 & 0.799 & 4.710 & 4.574  \\
Flex-Nano & 0.110 & 0.108 & 0.794 & 0.789 & 4.678 & 4.627  \\
Lite-Mono & 0.107 & 0.105 & 0.765 & 0.747 & 4.561 & 4.511 \\
ReForgeDepth (ours) & 0.110 & {0.104} & 0.894 & {0.746} & 4.748 & {4.521}  \\
\midrule
\multicolumn{7}{l}{\textit{(b) Teacher: Flex-Large}} \\
Lite-Mono-tiny & 0.110 & 0.105 & 0.837 & 0.773 & 4.710 & 4.516 \\
Flex-Nano & 0.110 & 0.106 & 0.794 & 0.757 & 4.678 & 4.547  \\
Lite-Mono-8M & 0.101 & 0.100 & 0.729 & 0.696 & 4.454 & 4.356  \\
ReForgeDepth (ours) & 0.110 & {0.102} & 0.894 & {0.726} & 4.748 & {4.459} \\
\bottomrule
\end{tabular}%
}\label{tb:woSS}
\end{table}

\begin{table}[ht]
\centering
\caption{\textbf{Sensitivity of the weight $\eta$ and its schedule on the KITTI Eigen split.} $\eta$ is the coefficient of the distillation loss in Eq.~\eqref{eq:total-loss}; ``Start ep.'' denotes the epoch at which distillation is activated, and ``Warm-up'' indicates whether $\eta$ increases gradually from zero.}
\label{tab:ablation_distill}   
\resizebox{0.5\textwidth}{!}{%
\begin{tabular}{cccc c}
\toprule
Weight $\eta$ & Warm-up & Start ep. & Teacher & Abs Rel$\downarrow$  \\
\midrule
\multicolumn{4}{l}{\textit{(a) Distillation weight, activated from epoch 0}} \\
0.1 & \XSolidBrush & 0 & LiteMono-8M  & 0.106  \\
0.1 & \XSolidBrush & 0 & Flex-Large  & 0.104  \\
0.05 & \XSolidBrush & 0 & LiteMono-8M & 0.107  \\
0.05 & \XSolidBrush & 0 & Flex-Large  & 0.105 \\
0.025 & \XSolidBrush & 0 & LiteMono-8M & 0.108 \\
0.025 & \XSolidBrush & 0 & Flex-Large  & 0.106  \\

\midrule
\multicolumn{4}{l}{\textit{(b) Distillation weight, activated from epoch 5}} \\
0.1 & \XSolidBrush & 5 & LiteMono-8M & 0.106  \\
0.1 & \XSolidBrush & 5 & Flex-Large  & 0.105  \\
0.05 & \XSolidBrush & 5 & LiteMono-8M & 0.107 \\
0.05 & \XSolidBrush & 5 & Flex-Large  & 0.105 \\
0.025 & \XSolidBrush & 5 & LiteMono-8M & 0.108 \\
0.025 & \XSolidBrush & 5 & Flex-Large  & 0.106  \\
\midrule
\multicolumn{4}{l}{\textit{(c) Warm-up ($\eta = 0.1$)}} \\
0.1 & \Checkmark & 5 & LiteMono-8M & 0.107\\
0.1 & \Checkmark & 5 & Flex-Large & 0.104  \\
0.1 & \Checkmark & 0 & LiteMono-8M & {0.104} \\
0.1 & \Checkmark & 0 & Flex-Large & {0.102}  \\
\bottomrule
\end{tabular}%
}
\end{table}

\begin{table}[ht]
\centering
\caption{\textbf{Ablation of training components and teacher choice.} Phase I denotes ImageNet initialization, Phase II denotes photometric self-supervised training, and Phase III denotes teacher distillation. A teacher name indicates that Phase III is used; \XSolidBrush indicates that it is omitted. Best Abs Rel is in \textbf{bold}.}
\label{tab:ablation_teacher}
\vspace{0.1cm}
\resizebox{0.5\textwidth}{!}{%
\begin{tabular}{lccccccccc}
\toprule
\multicolumn{3}{c}{Configuration} & \multicolumn{7}{c}{KITTI Eigen} \\
\cmidrule(lr){1-3} \cmidrule(lr){4-10}
 Phase I & Phase II & Phase III & Abs Rel$\downarrow$  \\
\midrule
\multicolumn{10}{l}{\textit{(a) Training-component ablation}} \\
 \XSolidBrush &\XSolidBrush  & Lite-Mono-8M & 0.120 \\
 \XSolidBrush &\XSolidBrush  & Flex-Large & 0.147  \\
   \XSolidBrush &\Checkmark& \XSolidBrush & 0.117  \\
  \XSolidBrush& \Checkmark & Lite-Mono-8M & 0.115 \\
\XSolidBrush& \Checkmark & Flex-Large & 0.115  \\
\Checkmark & \Checkmark  & \XSolidBrush &  0.110  \\
\Checkmark & \Checkmark  & Teacher-free (30 epoch.) &  0.110 \\
 \Checkmark & \XSolidBrush & Lite-Mono-8M & 0.113  \\
 \Checkmark & \XSolidBrush & Flex-Large & 0.111 & \\
\midrule
\multicolumn{10}{l}{\textit{(b) Teacher choice under the full SS strategy}} \\
 \Checkmark & \Checkmark & Depth Anything V3-Small & 0.114  \\
 \Checkmark & \Checkmark & Lite-Mono-3.1M & 0.107  \\
 \Checkmark & \Checkmark & Lite-Mono-8M & {0.104}  \\
 \Checkmark & \Checkmark & Flex-Large & \textbf{0.102}  \\
\bottomrule
\end{tabular}%
}
\end{table}

\subsection{Ablation Study}\label{sec:ablation}

\indent \textbf{DepthNet (See supplementary materials).} All variants are trained from scratch to isolate the architecture from pretraining. In every pairing, our GMD and DCGI match or outperform the Lite-Mono \cite{litemono} CDC and LGFI at roughly half the parameters and 25\% fewer FLOPs (0.124 at 1.29M vs.\ 2.67M); adding BR yields the best from-scratch result.

\indent \textbf{Signal Scaffolding (Tables~\ref{tb:woSS}, \ref{tab:ablation_distill}, \ref{tab:ablation_teacher}).} Table~\ref{tb:woSS} applies SS to different model scales under identical data. The gain decreases from 0.005 Abs Rel for Lite-Mono-tiny \cite{litemono} to 0.001 for Lite-Mono-8M \cite{litemono}, with the smallest model benefiting most. Under the same teacher, gains likewise shrink as model size grows. Teacher capability is thus a secondary factor; what dominates, instead, is likely the student's recoverable capacity and its ability to exploit the training signal \footnote{See Supplementary Materials for details.}. Table~\ref{tab:ablation_distill} shows that constant weights and delayed onset are suboptimal: full strength hinders optimization, while weaker weights lead to underfitting. Ramping the signal from zero performs best, allowing small networks to adapt gradually. Table \ref{tab:ablation_teacher} further shows that teacher quality and domain alignment matter: low-redundancy students amplify teacher noise, while cross-domain teachers introduce gaps that hinder learning. Simply extending training epochs provides little additional benefit.
\\
\indent Overall, SS has enhanced the depth accuracy of compact models without increasing the inference burden. The longer the self-supervised training time of the traditional training strategy, the more limited the improvement achieved. These results indicate that the traditional training strategy may not be able to achieve the performance that compact models can reach, while SS reduces the mismatch between training strategy and model capacity and enables small models to better utilize their existing expressive capabilities.

%% file: sec/5_Conclusion.tex
\section{Conclusion}
This paper reveals the mismatch between the training strategy and model capacity in lightweight self-supervised depth estimation. We propose ``Signal Scaffolding''---a three-phase strategy that gradually adjusts the training signals to approach the target task, thereby unlocking the latent representational capacity of the lightweight model without the need for additional dataset or parameters. In addition, we design ReForgeDepth, which reduces the number of parameters while retaining detailed information and the receptive field. On the KITTI and Cityscapes datasets, ReForgeDepth achieves the best balance of accuracy and efficiency among the compared methods, and outperforms other models of the same scale by a large margin.

Two {limitations} remain: photometric consistency degrades depth around moving objects, and the lack of ground-truth depth caps accuracy. Future work will integrate motion-aware masking into Signal Scaffolding and explore stronger geometric priors in even more compact architectures, narrowing the supervised gap while preserving on-device efficiency.

%% file: sec/LLM.tex
\section*{LLM Usage Statement}

During the preparation of this work, the authors used a large language model (LLM) to polish the language of the manuscript. The tool was not used to generate technical content, to conduct or analyze experiments, or to produce citations. All AI-assisted text was reviewed and edited by the authors, who take full responsibility for the content of this paper.
\newpage

%% file: sec/6_suppl.tex
\setcounter{page}{1}
\maketitlesupplementary

\section{Supplementary data and results}
\begin{figure*}[htbp] 
  \centering 
  \includegraphics[width=0.65\textwidth]{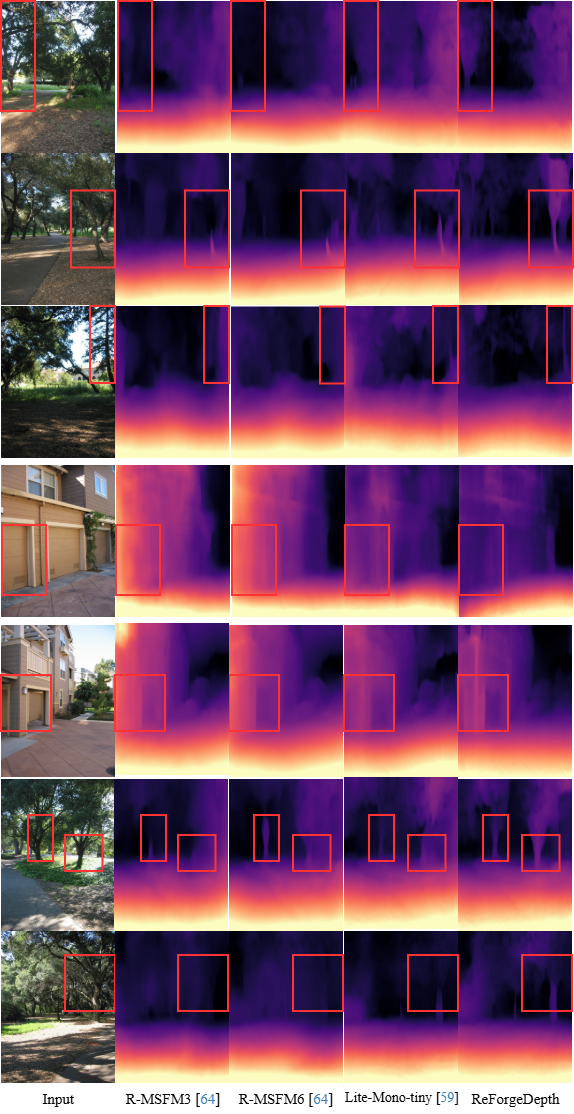} 
  \captionsetup{justification=centering}
\caption{\textbf{Qualitative results on the Make3D dataset.} Validation on challenging scenes from the Make3D dataset. Compared with prior lightweight models, ReForgeDepth captures a wider receptive field and finer details, demonstrating superior generalization to unseen data.}
  \label{fig:exampleMake3Dsm} 
\end{figure*}

This section provides the complete data and results that complement the main paper: full quantitative comparisons on KITTI \cite{kitti} and Cityscapes \cite{cordts2016cityscapes} (Tables \ref{tab:kittism} and \ref{tab:cityscapessm}), qualitative results on Make3D \cite{make3d} (Figure \ref{fig:exampleMake3Dsm}), and extended ablations on Signal Scaffolding, the distillation weight, the architecture, and the progressive training framework (Tables \ref{tb:woSSsm},  \ref{tab:ablation_distillsm}, \ref{tab:ablation_archsm} and \ref{tab:ablation_teachersm}).

Table~\ref{tab:kittism} reports the complete results on the KITTI Eigen split under three settings. 
\textbf{1024$\times$320 (M$^*$).} In models with fewer than 5M parameters, ReForgeDepth uses only 1.5M parameters and achieves the best or near-best results on all error metrics. Its absolute relative error ratio is as low as 0.001 compared to the large-capacity model Lite-Mono-8M \cite{litemono}, which is 5.8 times larger in size; at the same time, it outperforms most of the other high-capacity baseline models in terms of most metrics, including MViTDepth \cite{gao2023edge}, HR-Depth \cite{lyu2021hr}, and Monodepth2-Res18 \cite{godard2019digging}.
\\
\textbf{640$\times$192 (M).} ReForgeDepth achieves 0.102 Abs Rel with 1.5M parameters, ranking first among all lightweight methods and exceeding most high-capacity models, e.g., LEA-depth \cite{liu2026lea}, Mono-ViFI-Res18 \cite{liu2024mono}, and ADDepth \cite{zhang2025lightweight}; only Lite-Mono-8M \cite{litemono}, Flex-Medium \cite{zhu2026towards} and Flex-Large \cite{zhu2026towards}, which are 8.5$\times$ and 10.1$\times$ larger, obtain lower errors.
\\
\textbf{Without ImageNet pretraining (M$^\dagger$).} Although removing pretraining degrades all methods, ReForgeDepth still ties EdgeMono \cite{wang2025edgemono} for the best Abs Rel with 2.9$\times$ fewer parameters, and achieves the best RMSE, RMSE log, $\delta<1.25$, and $\delta<1.25^{2}$, indicating that our gains do not rely on stronger initialization. Table~\ref{tab:cityscapessm} lists the results on Cityscapes \cite{cordts2016cityscapes} at 416$\times$128. ReForgeDepth reaches 0.091 Abs Rel with 1.5M parameters, reducing the error of PuriLight \cite{chen2025purilight} by 8.1\% and that of the equally sized Flex-Nano \cite{zhu2026towards} by 15.0\%. It also outperforms all high-capacity baselines except Flex-Medium \cite{zhu2026towards} and Flex-Large \cite{zhu2026towards}, e.g., ProDepth \cite{woo2024prodepth}, ManyDepth2 \cite{zhou2025manydepth2}, and ManyDepth \cite{watson2021manydepth}.

Figure~\ref{fig:exampleMake3Dsm} shows qualitative predictions on challenging scenes from Make3D \cite{make3d}. The model is trained solely on KITTI \cite{kitti} and evaluated on Make3D \cite{make3d} without any fine-tuning \cite{godard2019digging}. ReForgeDepth predicts depth over a wider scene range and recovers finer structures than prior lightweight models, confirming its generalization beyond the training domain.

\begin{table*}[htbp]
\centering
\captionsetup{justification=centering}
\caption{\textbf{Quantitative results.} We compare ReForgeDepth with representative methods on the Cityscapes . The best and second-best results in lightweight are highlighted in \textbf{bold} and \underline{underlined}, respectively. M: monocular videos, MS: monocular videos + semantic segmentation.}
\footnotesize
\setlength{\tabcolsep}{3pt} 
\resizebox{0.9\textwidth}{!}{%
\begin{tabular}{ccccccccccc}
\toprule
\multirow{2}{*}{Method} & \multirow{2}{*}{Year} & \multirow{2}{*}{Train} &
\multicolumn{4}{c}{\textbf{Depth Error} $\downarrow$} &
\multicolumn{3}{c}{\textbf{Depth Accuracy} $\uparrow$} &
\multirow{2}{*}{\begin{tabular}[c]{@{}c@{}} \textbf{Model Size} $\downarrow$\\ Params.\end{tabular}} \\
 & & & Abs Rel & Sq Rel & RMSE & RMSE log & $\delta < 1.25$ & $\delta < 1.25^2$ & $\delta < 1.25^3$ & \\
\midrule
\midrule
\multicolumn{11}{c}{\textit{\textbf{416$\times$128 (M)}}} \\
\multicolumn{11}{c}{\textit{Lightweight ($<$5M)}} \\
\midrule

Flex-Nano \cite{zhu2026towards}  & 2026 & M & 0.107   & 1.261  & 6.133 & 0.164    & {0.893} &  0.971 & \underline{0.989} & \textbf{1.5M}   \\ 
PuriLight\cite{chen2025purilight} & 2025  & M & \underline{0.099}   & \underline{1.014}  & \underline{5.892}  & \underline{0.153}    & \underline{0.901} & \underline{0.976} & \textbf{0.992} & \underline{2.7M}   \\ 

Ours  & 2026  & M  & \textbf{0.091}   &\textbf{0.918}  & \textbf{5.590} & \textbf{0.146}    & \textbf{0.912} & \textbf{0.978}  & \textbf{0.992} & \textbf{1.5M}   \\

\midrule
\multicolumn{11}{c}{\textit{High-capacity }} \\
\midrule
ManyDepth  \cite{watson2021manydepth}  & 2021 & M & 0.114   & 1.193  & 6.223 & 0.170     & 0.875 & 0.967 & 0.989 & 29.3M  \\ 
ManyDepth2  \cite{zhou2025manydepth2}  & 2025& M & 0.097   & 0.792  & 5.827 & 0.154     & 0.903 & 0.975 & 0.993 & 22.6M  \\ 
DynamicDepth \cite{feng2022disentangling} & 2022 & MS & 0.103   & 1.000  & 5.867 & 0.157   & 0.895 &  0.974 & 0.991 & 29.3M  \\  
ProDepth \cite{woo2024prodepth}   & 2024 & M & 0.095   & 0.876  & 5.531 & 0.146     & 0.908 & 0.978 & 0.993 & 27.7M  \\ 
MonoViT \cite{zhao2022monovit}  & 2022  & M & 0.114   & 1.238  & 6.589 & 0.174   & 0.860 &  {0.965} & {0.990} & 27.9M  \\ 
FGTO~\cite{moon2024ground}  & 2024    & MS    & {0.096} & {0.930} & {5.806} & {0.152} & {0.905} & {0.976} & {0.992}  & {27.9M+}  \\ 
Self-Distillation~\cite{yu2025exploiting}  & 2025    & M    & {0.118} & {1.091} & {6.076} & {0.173} & {0.854} & {0.960} & {0.988}  & {27.9M+}  \\ 
Flex-Small \cite{zhu2026towards} & 2026    & M    & {0.100} & {1.078} & {5.813} & {0.153} & {0.904} & {0.975} & {0.991}  & {6.1M}  \\ 
Flex-Medium \cite{zhu2026towards} & 2026    & M    & {0.089} & {0.885} & {5.358} & {0.143} & {0.917} & {0.979} & {0.993}  & {12.7M}  \\ 
Flex-Large \cite{zhu2026towards} & 2026    & M    & {0.087} & {0.911} & {5.310} & {0.139} & {0.924} & {0.981} & {0.993}  & {15.2M}  \\

\bottomrule
\bottomrule
\end{tabular}%
}
\label{tab:cityscapessm}
\end{table*}

\textbf{Generality of Signal Scaffolding}
Table~\ref{tb:woSSsm} applies SS to four students under two teachers. SS improves every student--teacher pair, and the gain is inversely related to student size: Lite-Mono-tiny \cite{litemono} improves by 0.004--0.005, Lite-Mono \cite{litemono} by 0.002, and the 8.7M Lite-Mono-8M \cite{litemono} by only 0.001. This phenomenon can be quantified through a transitivity analysis. Let $A = Lite-Mono-tiny$, $B = Lite-Mono-8M$, and $C = Flex-Large$. If the teacher prior were the sole determinant, the gain would depend only on the teacher: since $C$ improves $B$ by 1\%, it should likewise improve $A$ by roughly 1\%; in practice, $A$ improves by 4.6\%---4.6 times the predicted value. Conversely, replacing student $A$'s teacher with the roughly 6\% stronger $C$ raises the gain only from 3.6\% to 4.6\% (+1.0\%), whereas swapping the student from $B$ to $A$ under the same teacher $C$ raises the gain from 1\% to 4.6\% (+3.6\%). Teacher capability is thus a secondary factor; what dominates, instead, is likely the student's recoverable capacity and its ability to exploit the training signal. Two independent observations are consistent with this reading. First, even after SS, $B$ still trails $C$ by 5\%, indicating that the teacher's knowledge is not transferred wholesale---$B$ absorbs only about one-sixth of the teacher-student gap, so the bottleneck lies in student capacity rather than the teacher prior. Second, under the same 1.5M parameter budget and the same teacher, ReForgeDepth improves far more than Flex-Nano \cite{zhu2026towards}, indicating that the recoverable capacity also depends on the architecture. These results confirm that SS, by reducing the capacity--training mismatch, lets small models better exploit their existing expressive power.

\begin{table*}[htbp]
\centering
\caption{\textbf{Generalization of the Signal Scaffolding (SS) on the KITTI Eigen split.} ``w/o SS'' denotes training with ImageNet initialization only, and ``w/ SS'' further applies Signal Scaffolding.}
\resizebox{0.85\textwidth}{!}{%
\begin{tabular}{l cc cc cc cc}
\toprule
& \multicolumn{2}{c}{Abs Rel$\downarrow$} & \multicolumn{2}{c}{Sq Rel$\downarrow$} & \multicolumn{2}{c}{RMSE$\downarrow$} & \multicolumn{2}{c}{RMSE log$\downarrow$} \\
\cmidrule(lr){2-3} \cmidrule(lr){4-5} \cmidrule(lr){6-7} \cmidrule(lr){8-9}
Method & w/o & w/ & w/o & w/ & w/o & w/ & w/o & w/ \\
\midrule
\multicolumn{9}{l}{\textit{(a) Teacher: Lite-Mono-8M \cite{litemono} }} \\
Lite-Mono-tiny \cite{litemono} & 0.110 & 0.106 & 0.837 & 0.799 & 4.710 & 4.574 & 0.187 & 0.182 \\
Flex-Nano \cite{zhu2026towards} & 0.110 & 0.108 & 0.794 & 0.789 & 4.678 & 4.627 & 0.184 & 0.184 \\
Lite-Mono \cite{litemono} & 0.107 & 0.105 & 0.765 & 0.747 & 4.561 & 4.511 & 0.183 & 0.181 \\
ReForgeDepth (ours) & 0.110 & {0.104} & 0.894 & {0.746} & 4.748 & {4.521} & 0.186 & {0.180} \\
\midrule
\multicolumn{9}{l}{\textit{(b) Teacher: Flex-Large \cite{zhu2026towards}}} \\
Lite-Mono-tiny \cite{litemono} & 0.110 & 0.105 & 0.837 & 0.773 & 4.710 & 4.516 & 0.187 & 0.180 \\
Flex-Nano \cite{zhu2026towards} & 0.110 & 0.106 & 0.794 & 0.757 & 4.678 & 4.547 & 0.184 & 0.182 \\
Lite-Mono-8M \cite{litemono} & 0.101 & 0.100 & 0.729 & 0.696 & 4.454 & 4.356 & 0.178 & 0.176 \\
ReForgeDepth (ours) & 0.110 & {0.102} & 0.894 & {0.726} & 4.748 & {4.459} & 0.186 & {0.179} \\
\bottomrule
\end{tabular}%
}\label{tb:woSSsm}
\end{table*}

\textbf{Sensitivity to the Distillation Weight $\eta$}
Table~\ref{tab:ablation_distillsm} examines $\eta$, its activation epoch, and warm-up. With a constant weight (a), $\eta=0.1$ is best and smaller weights underfit (0.107--0.108 vs.\ 0.106 for Lite-Mono-8M \cite{litemono}). Delaying the onset to epoch 5 (b) removes early guidance and slightly degrades the results. With warm-up (c), ramping $\eta$ from zero starting at epoch 0 is consistently the best across both teachers (0.104 and 0.102), supporting the design adopted in the main paper.

\textbf{Architecture Ablation}
Table~\ref{tab:ablation_archsm} compares our blocks with their Lite-Mono \cite{litemono} counterparts, all trained from scratch. Replacing CDC with GMD yields the same Abs Rel (0.124) with 51.7\% fewer parameters (1.29M vs.\ 2.67M) and 26.1\% fewer FLOPs. Replacing the CDC--LGFI pair with GMD--DCGI improves Abs Rel (0.121$\rightarrow$0.119) while reducing the model size from 3.07M to 1.53M.. Adding BR brings a final, consistent improvement on all metrics (0.119$\rightarrow$0.117) at a negligible cost of 0.12 GFLOPs. The resulting configuration coincides with the not-pretrained entry (0.117) in Table~\ref{tab:kittism}.

\textbf{Progressive Training and Teacher Selection}
Table~\ref{tab:ablation_teachersm} dissects the three-phase strategy and the teacher choice. (a)~Every phase is necessary: distilling directly from scratch collapses, especially with the strongest teacher Flex-Large \cite{zhu2026towards} (0.147); self-supervised Phase II alone reaches 0.117; adding ImageNet initialization lowers it to 0.110 and prolonging the training does not change it; and the full recipe with distillation reaches 0.104 (Lite-Mono-8M \cite{litemono}). Skipping Phase II while keeping Phase I and III also degrades, showing that the intermediate self-supervised phase is not redundant. (b)~Under the full strategy, the teacher's domain match and accuracy jointly determine the outcome: the cross-domain Depth Anything V3-Small \cite{lin2025dav3} even degrades the student, the weak in-domain Lite-Mono \cite{litemono} brings a limited gain, and stronger in-domain teachers improve the student monotonically, from 0.104 (Lite-Mono-8M \cite{litemono}) to 0.102 (Flex-Large \cite{zhu2026towards}).

\begin{table*}[htbp]
\centering
\captionsetup{justification=centering}
\caption{\textbf{Quantitative results.} We compare ReForgeDepth with representative methods on the KITTI Eigen split \cite{eig2015pr}. The best and second-best results in lightweight are highlighted in \textbf{bold} and \underline{underlined}, respectively. M$^\dagger$: not pre-trained on ImageNet \cite{imagenet}, M: KITTI monocular videos, MS: monocular videos + semantic segmentation, M$^*$: input resolution of $1024\times320$.}
\footnotesize
\renewcommand{\arraystretch}{0.84}
\setlength{\tabcolsep}{3pt} 
\resizebox{0.9\textwidth}{!}{%
\begin{tabular}{ccccccccccc}
\toprule
\multirow{2}{*}{Method} & \multirow{2}{*}{Year} & \multirow{2}{*}{Train} &
\multicolumn{4}{c}{\textbf{Depth Error} $\downarrow$} &
\multicolumn{3}{c}{\textbf{Depth Accuracy} $\uparrow$} &
\multirow{2}{*}{\begin{tabular}[c]{@{}c@{}} \textbf{Model Size} $\downarrow$\\ Params.\end{tabular}} \\
 & & & Abs Rel & Sq Rel & RMSE & RMSE log & $\delta < 1.25$ & $\delta < 1.25^2$ & $\delta < 1.25^3$ & \\
\midrule
\midrule
\multicolumn{11}{c}{\textit{\textbf{1024$\times$320 (M$^*$)}}} \\
\multicolumn{11}{c}{\textit{Lightweight ($<$5M)}} \\
\midrule

R-MSFM3 \cite{rmsfm} & 2021    & M$^*$     & 0.112   & 0.773  & 4.581 & 0.189     & 0.879 & 0.960 & 0.982 & 3.5M   \\  
R-MSFM6 \cite{rmsfm} & 2021    & M$^*$     & 0.108   & 0.748  & 4.470 & 0.185     & 0.889 & 0.963 & 0.982 & 3.8M   \\ 
MViTDepth-tiny \cite{gao2023edge}   & 2023    & M$^*$    & {0.109} & {0.818} & {4.603} & {0.184} & {0.887} & {0.963} & {0.983}  & \underline{1.8M}  \\ 
MViTDepth-small  \cite{gao2023edge}   & 2023    & M$^*$    & {0.104} & {0.772} & {4.460} & {0.180} & {0.893} & \underline{0.965} & \underline{0.984}  & {2.8M}  \\ 
Lite-Mono-tiny  \cite{litemono} & 2023   & M$^*$    & {0.104} & {0.764} & {4.487} & {0.180} & {0.892} & {0.964} & {0.983}  & {2.2M}  \\ 
Lite-Mono-small  \cite{litemono} & 2023    & M$^*$    & {0.103} & {0.757} & {4.449} & {0.180} & {0.894} & {0.964} & {0.983}  & {2.5M}  \\ 
Lite-Mono \cite{litemono} & 2023   & M$^*$    & {0.102} & {0.746} & {4.444} & {0.179} & {0.896} & \underline{0.965} & {0.983}  & {3.1M}  \\ 
LEDepth  \cite{zhang2025ledepth} & 2025   & M$^*$    & {0.101} & {0.744} & {4.425} & \underline{0.177} & {0.899} & \underline{0.965} & \textbf{0.985}  & {3.1M}  \\ 
PuriLight \cite{chen2025purilight}  & 2025 & M$^*$  & \underline{0.100}   & \underline{0.718}  & \underline{4.403} & \underline{0.177}    & \underline{0.901} & \textbf{0.967}  & \textbf{0.985} & {2.7M}   \\
Ours  & 2026  & M$^*$  & \textbf{0.098}   &\textbf{0.712}  & \textbf{4.339} & \textbf{0.174}    & \textbf{0.902} & \textbf{0.967}  & \underline{0.984} & \textbf{1.5M}   \\
\midrule
\multicolumn{11}{c}{\textit{High-capacity }} \\
\midrule
Monodepth2-Res18 \cite{godard2019digging} & 2019 & M$^*$ & 0.115   & 0.882  & 4.701 & 0.190     & 0.879 & 0.961 & 0.982 & 14.3M  \\ 
HR-Depth \cite{lyu2021hr} & 2021    & M$^*$     & 0.106   & 0.755  & 4.472 & 0.181     & 0.892 & {0.966} & {0.984} & 14.7M  \\ 
MViTDepth \cite{gao2023edge}    & 2023    & M$^*$    & {0.102} & {0.756} & {4.415} & {0.177} & {0.899} & {0.966} & {0.984}  & {6.3M}  \\ 
Lite-Mono-8M \cite{litemono} & 2023   & M$^*$    & {0.097} & {0.710} & {4.309} & {0.174} & {0.905} & {0.967} & {0.984}  & {8.7M}  \\ 
\midrule
\midrule
\multicolumn{11}{c}{\textit{\textbf{640$\times$192 (M)}}} \\
\multicolumn{11}{c}{\textit{Lightweight ($<$5M)}} \\
\midrule

Lite-Mono-tiny  \cite{litemono} & 2023    & M    & {0.110} & {0.837} & {4.710} & {0.187} & {0.880} & {0.960} & {0.982}  & {2.2M}  \\ 
Lite-Mono-small  \cite{litemono} & 2023    & M    & {0.110} & {0.802} & {4.671} & {0.186} & {0.879} & {0.961} & {0.982}  & {2.5M}  \\ 
Lite-Mono \cite{litemono}  & 2023    & M    & {0.107} & {0.765} & {4.561} & {0.183} & {0.886} & {0.963} & {0.983}  & {3.1M}  \\ 
MViTDepth-tiny \cite{gao2023edge}    & 2023    & M    & {0.111} & {0.825} & {4.731} & {0.187} & {0.881} & {0.961} & {0.983}  & \underline{1.8M}  \\ 
MViTDepth-small \cite{gao2023edge}    & 2023    & M    & {0.107} & {0.788} & {4.579} & {0.183} & {0.887} & {0.963} & {0.983}  & {2.8M}  \\ 
R-MSFM3-GC \cite{zhou2024recurrent}  & 2024 & M & 0.112   & 0.789  & 4.621 & 0.189    & 0.880 &  0.960 & 0.982 & 3.5M   \\ 
R-MSFM6-GC \cite{zhou2024recurrent}  & 2024  & M & 0.112   & 0.789  & 4.597  & 0.189    & 0.881 & 0.961 & 0.981 & 3.8M   \\ 
LEDepth \cite{zhang2025ledepth}  & 2025    & M    & \underline{0.106} & {0.780} & {4.602} & {0.183} &{0.888} & {0.963} & \textbf{0.985}  & {3.1M}  \\ 
ADDepth-tiny \cite{zhang2025lightweight}   & 2025    & M    & {0.112} & {0.815} & {4.642} & {0.186} & {0.880} & {0.962} & {0.983}  & \underline{1.8M}  \\ 
ADDepth-small \cite{zhang2025lightweight}   & 2025    & M    & {0.108} & \underline{0.741} & {4.503} & {0.182} & {0.885} & \textbf{0.965} & \underline{0.984}  & {2.8M}  \\ 
EdgeMono \cite{wang2025edgemono}   & 2025    & M    & {0.107} & {0.799} & {4.606} & {0.183} & {0.886} & {0.962} & {0.982}  & {4.3M}  \\ 
PuriLight \cite{chen2025purilight}  & 2025 & M  & \underline{0.106}   & {0.747}  & {4.536} & {0.182}    & \underline{0.890} & \underline{0.964}  & \underline{0.984} & {2.7M}   \\
LEA-depth-tiny \cite{liu2026lea}  & 2026    & M    & {0.109} & {0.776} & {4.528} & {0.182} & {0.884} & \underline{0.964} & {0.983}  & \underline{1.8M}  \\ 
LEA-depth-small \cite{liu2026lea}  & 2026    & M    & {0.107} & {0.758} & \underline{4.489} & \underline{0.181} & {0.888} & \underline{0.964} & \underline{0.984}  & {3.0M}  \\ 
Flex-Nano \cite{zhu2026towards} & 2026    & M    & {0.110} & {0.794} & {4.678} & {0.184} & {0.878} & {0.961} & {0.983}  & \textbf{1.5M}  \\ 
Ours & 2026  & M  & \textbf{0.102}   &\textbf{0.726}  & \textbf{4.459} & \textbf{0.179}    & \textbf{0.891} & \underline{0.964}  & {0.983} & \textbf{1.5M}   \\

\midrule
\multicolumn{11}{c}{\textit{High-capacity }} \\
\midrule
Monodepth2-Res18 \cite{godard2019digging}   & 2019 & M & 0.115   & 0.903  & 4.863 & 0.193     & 0.877 & 0.959 & 0.981 & 14.3M  \\ 
Monodepth2-Res50 \cite{godard2019digging}  & 2019 & M & 0.110   & 0.831  & 4.642 & 0.187   & 0.883 &  0.962 & 0.982 & 32.5M  \\  
SGDepth \cite{kli2020}  & 2020 & MS & 0.113   & 0.835  & 4.693 & 0.191     & 0.879 & 0.961 & 0.981 & 16.3M  \\ 
MonoFormer \cite{bae2022monoformer}  & 2022  & M & 0.108   & 0.806  & 4.594 & 0.184   & 0.884 &  {0.963} & {0.983} & 23.9M+  \\ 
MViTDepth \cite{gao2023edge}     & 2023    & M    & {0.106} & {0.766} & {4.507} & {0.180} & {0.891} & {0.964} & {0.984}  & {6.3M}  \\ 
Lite-Mono-8M \cite{litemono} & 2023   & M    & {0.101} & {0.729} & {4.454} & {0.178} & {0.897} & {0.965} & {0.983}  & {8.7M}  \\
Mono-ViFI-Res18 \cite{liu2024mono}   & 2024    & M    & {0.105} & {0.708} & {4.446} & {0.179} & {0.887} & {0.965} & {0.984}  & {14.3M}  \\ 
R-MSFMX3-GC \cite{zhou2024recurrent}  & 2024    & M    & {0.111} & {0.765} & {4.591} & {0.189} & {0.880} & {0.961} & {0.982}  & {5.0M}  \\ 
R-MSFMX6-GC \cite{zhou2024recurrent}  & 2024    & M    & {0.109} & {0.761} & {4.552} & {0.186} & {0.885} & {0.962} & {0.982}  & {5.2M}  \\ 
ADDepth \cite{zhang2025lightweight}   & 2025    & M    & {0.106} & {0.712} & {4.425} & {0.180} & {0.889} & {0.965} & {0.984}  & {6.3M}  \\ 
LEA-depth \cite{liu2026lea}  & 2026    & M    & {0.103} & {0.718} & {4.406} & {0.178} & {0.893} & {0.966} & {0.984}  & {5.8M}  \\ 
Flex-Small \cite{zhu2026towards} & 2026    & M    & {0.104} & {0.713} & {4.458} & {0.179} & {0.890} & {0.964} & {0.983}  & {6.1M}  \\ 
Flex-Medium \cite{zhu2026towards} & 2026    & M    & {0.096} & {0.639} & {4.253} & {0.172} & {0.903} & {0.968} & {0.985}  & {12.7M}  \\ 
Flex-Large \cite{zhu2026towards} & 2026    & M    & {0.095} & {0.642} & {4.199} & {0.171} & {0.906} & {0.968} & {0.984}  & {15.2M}  \\ 

\midrule
\midrule
\multicolumn{11}{c}{\textit{\textbf{640$\times$192, Not Pre-trained on ImageNet (M$^\dagger$)}}} \\
\multicolumn{11}{c}{\textit{Lightweight ($<$5M)}} \\
\midrule

R-MSFM3 \cite{rmsfm}     & 2021     & M$^\dagger$     & 0.128   & 0.965  & 5.019 & 0.207    & 0.853 &  0.951 & 0.977 & 3.5M   \\
R-MSFM6 \cite{rmsfm}     & 2021    & M$^\dagger$      & 0.126   & 0.944  & 4.981  & {0.204}    & 0.857 & 0.952 & 0.978 & 3.8M   \\
Lite-Mono-Tiny  \cite{litemono} & 2023    & M$^\dagger$    & {0.125} & {0.935} & {4.986} & {0.204} & {0.853} & {0.950} & {0.978}  & {2.2M}  \\ 
Lite-Mono-Small  \cite{litemono} & 2023    & M$^\dagger$    & {0.123} & {0.919} & {4.926} & {0.202} & {0.859} & {0.951} & {0.977}  & {2.5M}  \\ 
Lite-Mono \cite{litemono}    & 2023  & M$^\dagger$   & {0.121} & {0.876} & {4.918} & {0.199} & {0.859} & {0.953} & \underline{0.980}  & {3.1M}  \\ 
RTIA-Mono \cite{zhao2025rtia} & 2025    & M$^\dagger$    & \underline{0.118} & \textbf{0.858} & \underline{4.810} & \underline{0.196} & {0.865} & \underline{0.955} & \underline{0.980}  & \textbf{1.4M}  \\ 
EdgeMono \cite{wang2025edgemono}& 2025    & M$^\dagger$    & \textbf{0.117} & {0.886} & {4.817} & \textbf{0.195} & \underline{0.867} & \textbf{0.957} & \textbf{0.981}  & {4.3M}  \\ 
PuriLight \cite{chen2025purilight}   & 2025     & M$^\dagger$       &{0.121}   & {0.908}  & {4.908} & {0.199}    & {0.862} & {0.954}  & \underline{0.980} & {2.7M}   \\
Ours  & 2026   & M$^\dagger$   & \textbf{0.117} & \underline{0.860} & \textbf{4.767} & \textbf{0.195} & \textbf{0.868} & \textbf{0.957} & \underline{0.980} &  \underline{1.5M} \\

\midrule
\multicolumn{11}{c}{\textit{High-capacity }} \\
\midrule
Monodepth2-Res18 \cite{godard2019digging}  & 2019  & M$^\dagger$ & 0.132   & 1.044  & 5.142 & 0.210     & 0.845 & 0.948 & 0.977 & 14.3M  \\
Monodepth2-Res50 \cite{godard2019digging}  & 2019  & M$^\dagger$ & 0.131   & 1.023  & 5.046 & 0.206   & 0.849 &  0.951 & {0.979} & 32.5M  \\

\bottomrule
\bottomrule
\end{tabular}%
}
\label{tab:kittism}
\end{table*}

\begin{table*}[htbp]
\centering
\caption{\textbf{Sensitivity of the weight $\eta$ and its schedule on the KITTI Eigen split.} $\eta$ is the coefficient of the distillation loss in Eq.~\eqref{eq:total-loss}; ``Start ep.'' denotes the epoch at which distillation is activated, and ``Warm-up'' indicates whether $\eta$ increases gradually from zero.}
\label{tab:ablation_distillsm}   
\resizebox{0.85\textwidth}{!}{%
\begin{tabular}{cccc ccccccc}
\toprule
Weight $\eta$ & Warm-up & Start ep. & Teacher & Abs Rel$\downarrow$ & Sq Rel$\downarrow$ & RMSE$\downarrow$ & RMSE log$\downarrow$ & $\delta<1.25\uparrow$ & $\delta<1.25^{2}\uparrow$ & $\delta<1.25^{3}\uparrow$ \\
\midrule
\multicolumn{10}{l}{\textit{(a) Distillation weight, activated from epoch 0}} \\
0.1 & \XSolidBrush & 0 & LiteMono-8M \cite{litemono}  & 0.106 & 0.795 & 4.603 & 0.183 & 0.887 & 0.962 & 0.983 \\
0.1 & \XSolidBrush & 0 & Flex-Large \cite{zhu2026towards}  & 0.104 & 0.772 & 4.552 & 0.181 & 0.888 & 0.963 & 0.983 \\
0.05 & \XSolidBrush & 0 & LiteMono-8M \cite{litemono} & 0.107 & 0.820 & 4.641 & 0.184 & 0.886 & 0.962 & 0.982 \\
0.05 & \XSolidBrush & 0 & Flex-Large \cite{zhu2026towards} & 0.105 & 0.795 & 4.605 & 0.182 & 0.887 & 0.962 & 0.983 \\
0.025 & \XSolidBrush & 0 & LiteMono-8M \cite{litemono} & 0.108 & 0.841 & 4.675 & 0.184 & 0.885 & 0.962 & 0.982 \\
0.025 & \XSolidBrush & 0 & Flex-Large \cite{zhu2026towards}  & 0.106 & 0.804 & 4.625 & 0.183 & 0.886 & 0.962 & 0.983 \\

\midrule
\multicolumn{10}{l}{\textit{(b) Distillation weight, activated from epoch 5}} \\
0.1 & \XSolidBrush & 5 & LiteMono-8M \cite{litemono} & 0.106 & 0.810 & 4.628 & 0.183 & 0.886 & 0.962 & 0.982 \\
0.1 & \XSolidBrush & 5 & Flex-Large \cite{zhu2026towards}  & 0.105 & 0.773 & 4.567 & 0.181 & 0.888 & 0.963 & 0.983 \\
0.05 & \XSolidBrush & 5 & LiteMono-8M \cite{litemono} & 0.107 & 0.814 & 4.648 & 0.184 & 0.886 & 0.962 & 0.982 \\
0.05 & \XSolidBrush & 5 & Flex-Large \cite{zhu2026towards}  & 0.105 & 0.788 & 4.607 & 0.183 & 0.887 & 0.962 & 0.983 \\
0.025 & \XSolidBrush & 5 & LiteMono-8M \cite{litemono} & 0.108 & 0.836 & 4.673 & 0.184 & 0.886 & 0.962 & 0.982 \\
0.025 & \XSolidBrush & 5 & Flex-Large \cite{zhu2026towards}  & 0.106 & 0.803 & 4.628 & 0.183 & 0.887 & 0.962 & 0.982 \\
\midrule
\multicolumn{10}{l}{\textit{(c) Warm-up ($\eta = 0.1$)}} \\
0.1 & \Checkmark & 5 & LiteMono-8M \cite{litemono} & 0.107 & 0.812 & 4.631 & 0.183 & 0.887 & 0.962 & 0.982 \\
0.1 & \Checkmark & 5 & Flex-Large \cite{zhu2026towards} & 0.104 & 0.769 & 4.565 & 0.181 & 0.888 & 0.963 & 0.983 \\
0.1 & \Checkmark & 0 & LiteMono-8M \cite{litemono} & {0.104} & {0.746} & {4.521} & {0.180} & {0.887} & {0.964} & {0.983} \\
0.1 & \Checkmark & 0 & Flex-Large \cite{zhu2026towards} & {0.102} & {0.726} & {4.459} & {0.179} & {0.891} & {0.964} & {0.983} \\
\bottomrule
\end{tabular}%
}
\end{table*}

\begin{table*}[t]
\centering
\caption{\textbf{Ablation study of the ReForgeDepth architecture on the KITTI Eigen split.} CDC and LGFI are the encoder blocks and the feature-interaction module of Lite-Mono, respectively; \textbf{GMD}, \textbf{DCGI}, and \textbf{BR} are our contributions. All experiments were trained from scratch to highlight the impact of the architecture. Best accuracy is in \textbf{bold}.}
\label{tab:ablation_archsm}
\vspace{0.1cm}
\resizebox{0.95\textwidth}{!}{%
\begin{tabular}{ccccc c ccccccc cc}
\toprule
\multicolumn{5}{c}{Encoder configuration} & \multicolumn{7}{c}{Accuracy} & \multicolumn{2}{c}{Complexity} \\
\cmidrule(lr){1-5} \cmidrule(lr){6-12} \cmidrule(lr){13-14}
CDC & \textbf{GMD} & LGFI & \textbf{DCGI} & \textbf{BR} & Abs Rel$\downarrow$ & Sq Rel$\downarrow$ & RMSE$\downarrow$ & RMSE log$\downarrow$ & $\delta<1.25\uparrow$ & $\delta<1.25^{2}\uparrow$ & $\delta<1.25^{3}\uparrow$ & Params (M)$\downarrow$ & FLOPs (G)$\downarrow$ \\
\midrule
\Checkmark & & & & & 0.124 & 0.995 & 4.995 & 0.199 & 0.860 & 0.954 & 0.979 & 2.67 & 4.403 \\
& \Checkmark & & & & 0.124 & 0.952 & 4.983 & 0.203 & 0.850 & 0.952 & 0.979 & \textbf{1.29} & \textbf{3.253} \\
\Checkmark & & \Checkmark & & & 0.121 & 0.876 & 4.918 & 0.199 & 0.859 & 0.953 & 0.980 & 3.07 & 5.032 \\
& \Checkmark & \Checkmark & & & 0.120 & 0.880 & 4.874 & 0.199 & 0.861 & 0.953 & 0.979 & 1.70 & 3.760 \\
\Checkmark & & & \Checkmark & & 0.121 & 0.942 & 4.872 & 0.199 & 0.864 & 0.954 & 0.979 & 2.90 & 4.775 \\
& \Checkmark & & \Checkmark & & 0.119 & 0.880 & 4.831 & 0.196 & 0.866 & 0.956 & 0.981 & 1.53 & 3.503 \\
\midrule
& \Checkmark & & \Checkmark & \Checkmark & \textbf{0.117} & \textbf{0.860} & \textbf{4.767} & \textbf{0.195} & \textbf{0.868} & \textbf{0.957} & \textbf{0.980} & 1.53 & 3.626 \\
\bottomrule
\end{tabular}%
}
\end{table*}

\begin{table*}[htbp]
\centering
\caption{\textbf{Ablation of training components and teacher choice.} Phase I denotes ImageNet initialization, Phase II denotes photometric self-supervised training, and Phase III denotes teacher distillation. A teacher name indicates that Phase III is used; \XSolidBrush indicates that it is omitted. Best Abs Rel is in \textbf{bold}.}
\label{tab:ablation_teachersm}
\vspace{0.1cm}
\resizebox{0.85\textwidth}{!}{%
\begin{tabular}{lccccccccc}
\toprule
\multicolumn{3}{c}{Configuration} & \multicolumn{7}{c}{KITTI Eigen} \\
\cmidrule(lr){1-3} \cmidrule(lr){4-10}
 Phase I & Phase II & Phase III & Abs Rel$\downarrow$ & Sq Rel$\downarrow$ & RMSE$\downarrow$ & RMSE log$\downarrow$ & $\delta<1.25\uparrow$ & $\delta<1.25^{2}\uparrow$ & $\delta<1.25^{3}\uparrow$ \\
\midrule
\multicolumn{10}{l}{\textit{(a) Training strategies}} \\
 \XSolidBrush &\XSolidBrush  & Lite-Mono-8M \cite{litemono} & 0.120 & 0.890 & 4.887 & 0.198 & 0.860 & 0.954 & 0.980 \\
 \XSolidBrush &\XSolidBrush  & Flex-Large \cite{zhu2026towards} & 0.147 & 1.105 & 5.444 & 0.219 & 0.807 & 0.939 & 0.977 \\
   \XSolidBrush &\Checkmark& \XSolidBrush & 0.117 & 0.860 & 4.767 & 0.195 & 0.868 & 0.957 & 0.980 \\
  \XSolidBrush& \Checkmark & Lite-Mono-8M \cite{litemono} & 0.115 & 0.833 & 4.739 & 0.193 & 0.871 & 0.958 & 0.981 \\
\XSolidBrush& \Checkmark & Flex-Large \cite{zhu2026towards} & 0.115 & 0.836 & 4.722 & 0.193 & 0.872 & 0.958 & 0.981 \\
\Checkmark & \Checkmark  & \XSolidBrush &  0.110 & 0.894 & 4.748 & 0.186 & 0.884 & 0.961 & 0.982 \\
\Checkmark & \Checkmark  & Teacher-free (30 epoch.) &  0.110 & 0.871 & 4.714 & 0.186 & 0.884 & 0.961 & 0.982 \\
 \Checkmark & \XSolidBrush & Lite-Mono-8M \cite{litemono} & 0.113 & 0.847 & 4.642 & 0.185 & 0.878 & 0.962 & 0.983 \\
 \Checkmark & \XSolidBrush & Flex-Large \cite{zhu2026towards} & 0.111 & 0.762 & 4.636 & 0.185 & 0.873 & 0.961 & {0.984} \\
\midrule
\multicolumn{10}{l}{\textit{(b) Teacher model }} \\
 \Checkmark & \Checkmark & Depth Anything V3-Small \cite{lin2025dav3} & 0.114 & 0.875 & 5.642 & 0.205 & 0.848 & 0.945 & 0.978 \\
 \Checkmark & \Checkmark & Lite-Mono-3.1M \cite{litemono} & 0.107 & 0.810 & 4.588 & 0.184 & 0.884 & 0.962 & 0.982 \\
 \Checkmark & \Checkmark & Lite-Mono-8M \cite{litemono} & {0.104} & {0.746} & {4.521} & {0.180} & {0.887} & {0.964} & {0.983} \\
 \Checkmark & \Checkmark & Flex-Large \cite{zhu2026towards} & {0.102}   &{0.726}  & {4.459} & {0.179}    & {0.891} & {0.964}  & {0.983}  \\
\bottomrule
\end{tabular}%
}
\end{table*}